\documentclass[runningheads]{llncs}
\usepackage{graphicx}
\usepackage{comment}
\usepackage{booktabs}
\usepackage{xcolor}
\usepackage{soul}
\usepackage{multirow}
\usepackage{subcaption}
\usepackage{enumitem}% http://ctan.org/pkg/enumitem

\begin{document}

\title{Textual Entailment for Effective Triple Validation in Object Prediction}

%
%\titlerunning{Abbreviated paper title}
% If the paper title is too long for the running head, you can set
% an abbreviated paper title here
%
\author{Andrés García-Silva \and
Cristian Berrío\and
José Manuel Gómez-Pérez}
\authorrunning{A. Garcia-Silva et al.}
% First names are abbreviated in the running head.
% If there are more than two authors, 'et al.' is used.
%
\institute{Expert.ai, Language Technology Research Lab,\\ Poeta Joan Maragall 3, 28020 Madrid, Spain \\
%Springer Heidelberg, Tiergartenstr. 17, 69121 Heidelberg, Germany
\email{agarcia@expert.ai, cberrio@expert.ai, jmgomez@expert.ai}\\
\url{https://www.expert.ai} %\and
%ABC Institute, Rupert-Karls-University Heidelberg, Heidelberg, Germany\\
%\email{\{abc,lncs\}@uni-heidelberg.de}
}
\maketitle              % typeset the header of the contribution
\begin{abstract}

Knowledge base population seeks to expand knowledge graphs with facts that are typically extracted from a text corpus. Recently, language models pretrained on large corpora have been shown to contain factual knowledge that can be retrieved using cloze-style strategies. Such approach enables zero-shot recall of facts, showing competitive results in object prediction compared to supervised baselines. However, prompt-based fact retrieval can be brittle and heavily depend on the prompts and context used, which may produce results that are unintended or hallucinatory. We propose to use textual entailment to validate facts extracted from language models through cloze statements. Our results show that triple validation based on textual entailment improves language model predictions in different training regimes. Furthermore, we show that entailment-based triple validation is also effective to validate candidate facts extracted from other sources including existing knowledge graphs and text passages where named entities are recognized.

%Knowledge base population seeks to expand knowledge graphs with facts that are typically extracted from a text corpus. Recently, language models pretrained on large corpora have shown to contain factual knowledge that can be retrieved using cloze-style strategies. Such approach enables zero-shot recall of facts, showing competitive results in object prediction compared to supervised baselines. However, prompt-based fact retrieval can be brittle and heavily depend on the prompts and context used, which may produce results that are unintended or hallucinatory. We propose to use textual entailment to validate facts extracted from language models through cloze statements. Our results show that triple validation using textual entailment improves language models predictions in different training regimes. Furthermore, we show that triple validation is effective to extract facts from other sources including existing knowledge bases and text passages where named entities are recognized. 

\end{abstract}
\section{Introduction}
Knowledge Graphs arrange entities and relationships in a graph structure to represent knowledge \cite{speer2017conceptnet,miller1995wordnet}.
%general knowledge \cite{vrandevcic2014wikidata}, commonsense \cite{speer2017conceptnet}, linguistic \cite{miller1995wordnet} or domain-specific knowledge.
The edges of the graph describe relations between subject and object entities that are encoded as $<$\textit{subject relation object}$>$ triples. 
Knowledge graphs have applications in many areas, including search\footnote{https://blog.google/products/search/introducing-knowledge-graph-things-not/}, recommendation, and natural language processing~\cite{JiKGSurvey2022}. Nowadays the collaboration between editors and bots to curate and extend knowledge graphs has become common \cite{vrandevcic2014wikidata}. However, this is a complex and never-ending task and as a consequence, knowledge graphs are often incomplete \cite{west_knowledge_2014,Galagarra2017}. 

Knowledge Base Completion KBC \cite{balazevic-etal-2019-tucker} aims at predicting relations between existing entities. Similarly, the goal of the Knowledge Base Population KBP task \cite{ji_knowledge_2011} is to expand knowledge graphs with new facts discovered from text corpora. While in recent years a plethora of embeddings-based approaches have emerged for KBC \cite{JiKGSurvey2022}, KBP research %is stalled. 
has not progressed at the same speed due to the complexity of the pipelines \cite{ji_knowledge_2011,getman_laying_2018} and the lack of established benchmarks.\footnote{The KBP evaluation track of the TAC \cite{getman_laying_2018} is a long running initiative. However, manual system evaluation makes it hard to reproduce evaluation for new systems.}
%\footnote{Although some authors use KBC to refer to KBP \cite{west_knowledge_2014}, they are different processes. in KBC the objects are known in inference while in KBP they are unknown.}. 
% Common KBP pipelines \cite{ji_knowledge_2011} include supervised entity linking \cite{wu2020scalable}, and slot filling \cite{Huang2017ImprovingSF} components that need to be trained. More elaborated systems extract entities, relations, sentiments, and events from a text corpus to populate knowledge bases from scratch \cite{getman_laying_2018}. %However, training data, the main input of supervised learning, is scarce in real-life projects%targeting the population of knowledge graphs for specific domains
%, particularly for domain-specific knowledge graphs. %The generation of annotated data is heavy-cost and skilled labour demanding. Therefore a realistic scenario for KBP should include zero-shot and few-shot settings where no training data or just few data is available. 

Recently, language models have been revealed as promising resources for KBP. Language models trained on large text corpora encode different types of knowledge including syntactic \cite{liu_linguistic_2019}, semantic \cite{Teney2019LearnFromCon}, commonsense \cite{Zhou2020EvaluatingCommonsense}, and factual knowledge \cite{petroni_language_2019}. To elicit facts from the internal memory of a language model, researchers typically use cloze statements to make the language model fill in the masked tokens, e.g., John Lennon plays $<$MASK$>$. Cloze statements, also known as prompts in this context, enable zero-shot fact retrieval without any fine-tuning \cite{petroni_language_2019}. 
%Another line of work use probing classifiers, that are learned by fine-tuning the language model, to evaluate the knowledge encoded in language models \cite{Teney2019LearnFromCon}. 
Nevertheless, the knowledge encoded in the language model is limited to the data it has seen during pretraining. Additionally, prompt-based fact retrieval can be brittle~\cite{poerner2020bert} and heavily depend on the prompts and context used~\cite{cao-etal-2021-knowledgeable}, which may produce results that are unintended or hallucinatory. %Fine-tuning on the other hand cause that relational knowledge is forgotten in the process \cite{singh-etal-2020-bertnesia}.
For example, as a response to the previous prompt, BERT %\footnote{\url{https://huggingface.co/bert-base-uncased}}
would return {\tt guitar, piano, drums, himself, harmonica}. Lennon played percussion overdubs on some tracks, but he never actually played the drums. Further, while all of the remaining statements are true, "John Lennon plays himself" relates to his acting side, while we are interested in musical instruments. An apparently more specific prompt like John Lennon plays instrument $<$MASK$>$ returns {\tt here, there, too, himself, onstage}, adding even more noise. %Finally, neither prompted the saxophone, which he was known to play in some of his songs like "I'm Only Sleeping".

\begin{figure*}[ht!]
\includegraphics[width=\textwidth]{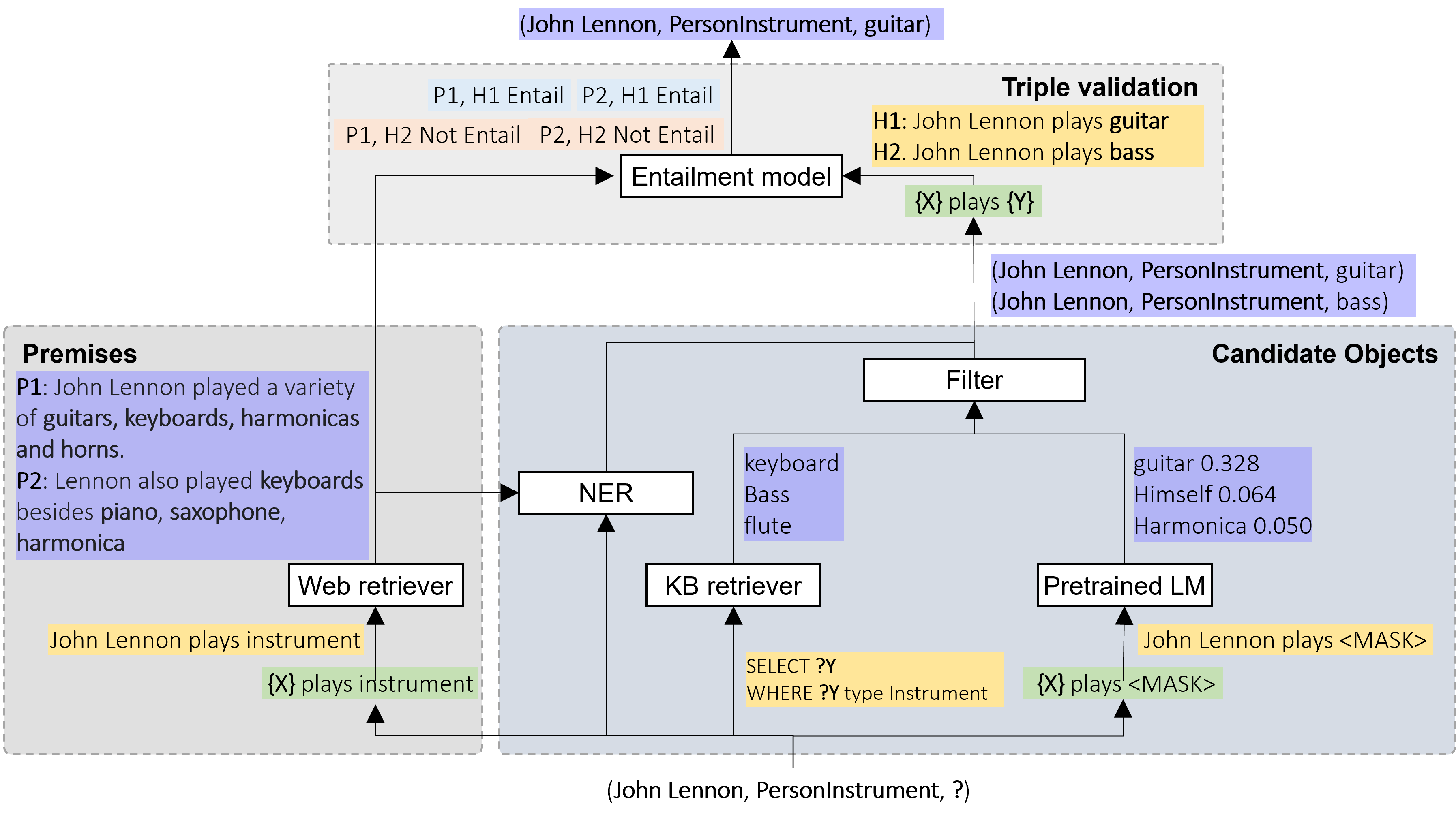}
\caption{\small SATORI architecture exemplified using as input pair \textit{John Lennon} in the subject and \textit{PersonInstrument} in the  relation.} \label{fig1}
\end{figure*}

To address such limitations we propose to validate candidate triples using textual entailment~\cite{richardson_probing_2020} against evidence retrieved from the Web. Within KBP, we focus on the object prediction task~\cite{singhania_lm-kbc_2022}. Given a subject entity and a relation, the goal is to predict every object that renders a valid triple, where such objects may not have been contained in the knowledge graph yet. %Object prediction is related to slot filling \cite{ji_knowledge_2011} in KBP, although it does not required previously annotated entities in the text. 
In this paper we present our system SATORI (Seek And enTail for Object pRedIction). As shown in Fig.~\ref{fig1}, SATORI obtains candidate objects from language models using cloze statements and generates candidate triples. To improve recall SATORI also considers other sources of candidate objects including external knowledge bases and named entities recognized in relevant text passages. A language model fine-tuned on the entailment task is used to validate whether the generated triples can be entailed from passages retrieved from the web. The objects of the triples validated by the model as entailment are the output of the system.  

%Instead of using language models only as source of knowledge for KBP, we also use them to infer whether facts can be drawn from text passages retrieved from the web. Language models can gain reasoning skills when they are fine-tuned on the entailment task \cite{richardson_probing_2020}. Our work is focused on object prediction, a particular task within KBP%, in zero-shot and few-shot training regime. 
%That is, given a subject entity and a relation the task goal is to predict every object, not necessarily already part of the knowledge graph, that make up valid triples. Object prediction is related to slot filling \cite{ji_knowledge_2011}, although it does not required previously annotated entities in the text. 

%Templates are the only prerequisite of SATORI. They are used to 
SATORI relies on templates to convert the input subject and relation pair into search engine queries to retrieve text passages, language model prompts to get candidate objects, and hypotheses describing candidate triples. Templates need to be defined only once per relation, and in its most basic form a template can be re-used across all the system components. For example, given the input pair \textit{(John Lennon, PersonIntrument)} and a relation template \{X\} \textit{plays}, we can submit the query \textit{John Lennon plays} to the search engine, prompt the language model with \textit{John Lennon plays $<$MASK$>$} for a candidate object, e.g., \textit{Guitar}, and validate the entailment of the hypothesis \textit{John Lennon plays Guitar} against the premises retrieved through our web search. Using language models as a source of factual knowledge, SATORI could also leverage breakthroughs in prompting and knowledge-based enhancement of language models~\cite{poerner2020bert}. 

Validating candidate triples generated from objects predicted using language models raises the following main research questions:
\begin{itemize}[noitemsep,topsep=0pt]
    \item Q1: Does candidate triple validation through textual entailment improve object prediction results over prompting a pretrained language model?
%    \item Q2: What makes textual entailment a valid proxy task for triple validation in object prediction?
    \item Q2: How do further pretraining the language model and fine-tuning for entailment-based triple validation under different data regimes impact object prediction?
    %\item Q3: What features of entailment models are important when it comes to validate triples?
    \item Q3: How do language models compare to other potential sources of structured knowledge, as well as to methods based on extracting information from text, for the generation of candidate triples?
\end{itemize}

In this paper, we investigate such research questions and present the following contributions. First, an approach for object prediction including the validation of triples using textual entailment. Second, an experimental study where different configurations of SATORI and baseline systems are evaluated using a gold dataset for object prediction \cite{singhania_lm-kbc_2022}. In such experiments we show that triple validation through textual entailment improves the performance of facts extracted from pretrained language models and also when additional training data is available. Finally, we compare language models with other sources including structured knowledge and unstructured text, which we process using extractive techniques repurposed for object prediction.

\section{Related Work}

\textbf{Knowledge base population.} %KBP goal is to %discover novel entities in a corpus, link them to corresponding knowledge base entries, then discover relation among the entities, and finally 
%expand a knowledge base with new facts \cite{ji_knowledge_2011} extracted from a corpus. 
A prototypical KBP architecture \cite{balog_populating_2018,ji_knowledge_2011} applies \textit{entity linking} \cite{wu2020scalable} on a corpus to recognize the entities mentioned in the text. Then a \textit{slot filling} process \cite{surdeanu2014slotfillingtac14} predicts an object, either an entity or a value for each subject entity and named relation, given the text where the entity appears. A key component of slot filling is a \textit{relation extraction} model \cite{adel2019type,Huang2017ImprovingSF} that identifies the relation between the entity and the candidate object. %The KBP track of the Text Analysis Conference \cite{ji_knowledge_2011,surdeanu2014slotfillingtac14} evaluated entity linking and slot filling separately, on a limited set of subject types restricted to person and organization. 
Thus, in KPB object prediction could complement the slot filling task since both share the goal of getting entities or values for a given subject and predicate. However, while slot filling leverages a large corpus to get the objects, in object prediction we rely on a language model to get objects and neither entity linking, relation extraction nor a text corpus are required. 

Similar to our work West et al. \cite{west_knowledge_2014} extract values from web search snippets although using question-answering (QA). Nevertheless, they experiment only with relations of subject type PERSON, and their approach only predict entities already in the knowledge graph. Unlike West et al., our work is framed in open world KBP \cite{shi_open-world_2018}, where predicted entities are not restricted to existing entities. %in the knowledge graph. %In fact in zero-shot KBP objects to be predicted are unknown to our model, and in few-shot only some of them have been seen during training. 

\textbf{Knowledge base completion.} KBC \cite{JiKGSurvey2022}  mostly refers to \textit{relation prediction} \cite{balazevic-etal-2019-tucker,zha_inductive_2022} where the goal is to recognize a relation between two existing entities. However, KBC de facto evaluation turns relation prediction into a ranking task where triples with right objects (or subjects) are expected at the top \cite{bordes_translating_2013,yao2019kg}. Such evaluation could lead to confusing KBC goal with object prediction. However in KBC the predicted objects are already part of the knowledge graph, while in object prediction they are not necessarily known. 

%Although some authors use the term \textit{Knowledge Base Completion} KBC to refer to KBP processes \cite{west_knowledge_2014}, KBC is mostly used in the literature for \textit{relation prediction} \cite{balazevic-etal-2019-tucker,zha_inductive_2022} to recognize a named relation between two entities in a knowledge graph. KBP and relation prediction are different tasks since in relation prediction the object entities are an input in inference time while in KBP they are unknown. 

\textbf{Prompts and entity-enhanced languages models.}
To elicit relational knowledge from language models Petroni et al. \cite{petroni_language_2019} use prompting. They show that prompting BERT achieves competitive results againts non-neural and supervised alternatives. Prompts can be hand-crafted  \cite{petroni_language_2019}, mined from a large corpus \cite{bouraoui2020inducing} or learned \cite{qin-eisner-2021-learning,shin_autoprompt_2020}. Li et al. \cite{Li2022TaskspecificPA} decompose prompts to split the task into multiple steps, and use task-specific pretraining for object prediction. Similar to our work Alivanistos et al. \cite{AlivanistosPromptProbing2022} add a fact probing step. However rather than using entailment for the validation, they ask a generative language model whether the generated fact is correct. Other works such as KnowBERT \cite{peters2019knowledge}, and E-BERT \cite{poerner2020bert} enhance BERT with pretrained entity embeddings, showing that such enhanced versions improve BERT results on LAMA \cite{petroni_language_2019}. 

\textbf{Textual Entailment.} %Also known as natural language inference 
It aims at recognizing, given two text fragments, whether the meaning of one text can be inferred (entailed) from the other \cite{dagan2005pascal}. 
%is as a directional relationship between a premise P and a hypothesis H, so that P entails H if a human reading P would infer that H is most likely true \cite{dagan2005pascal}. 
The state of the art for textual entailment \cite{wang_entailment_2021} is to fine-tune language models on datasets such as SNLI \cite{bowman-etal-2015-large} or MNLI \cite{williams-etal-2018-broad}. Researchers have reformulated several task as entailment. %for zero-shot and few-shot scenarios. 
Wang et al. \cite{wang_entailment_2021} convert single-sentence and sentence-pair classification tasks into entailment style. %and demonstrate that standard pretrained language models are effective few-shot learners.  
Entailment has been also used for zero-shot and few-shot relation extraction \cite{Sainz2021LabelVA}, answer validation \cite{rodrigo2009}, event argument extraction \cite{sainz_textual_2022}, and claim verification in fact checking \cite{guo-etal-2022-survey}. Preliminary experiments suggest that a slot filling validation filter using textual entailment could be useful for KBP systems \cite{Bentivogli2011TheSP}. 
To the best of our knowledge we are the first to use textual entailment to validate object prediction results.  

\textbf{Fact checking.} Fact checking \cite{guo-etal-2022-survey}, originally proposed for assessing whether claims made in written or spoken language are true, have been applied also to check the trustworthiness of facts in knowledge graphs. Approaches for truth scoring of facts can be broadly classified into two types \cite{kim-choi-2020-unsupervised}: approaches that use unstructured textual data to find supporting evidential sentences for a given statement \cite{gerber2015,syed2018,thorne-vlachos-2018-automated}, and  approaches that use a knowledge graph to find supporting evidential paths for a given statement \cite{shiralkar2017,syed2019}. Fact checking, is a step beyond our triple validation approach, where trustworthiness of the triples and the sources is evaluated. Trustworthiness analysis is out of the scope of out work that is defined in the research questions that we pose.

\section{Object Prediction}\label{SATORI}

\subsection{Task definition}
Let $F\subset E\times R\times E$ be the set of facts in a knowledge graph, where E is the set of entities and R the set of relations.
%A knowledge graph can be formally defined as $G=(E,R,F)$ where E is the set of entities, R the set of relations, and $F\subset E\times R\times E$ is the set of facts. 
Given $s \in E$ and $r \in R$, the goal of the object prediction task is to find every $o \in E \cup E'$ where $E'$ is the complement of $E$ such that the triple $(s,r,o)$ is a valid fact. Valid triples are added to $F$, and $E$ is expanded when $o$ is not an existing entity in the graph. %In zero-shot prediction the KBP system has not seen any example of object prediction data, or in other words any triple $(s,r,o)$, during training, while in few-shot prediction the system has available a few examples to adjust its internal model during training. 
In object prediction, as in other KBP tasks, the knowledge graph schema is available. The schema defines the relations in the knowledge graph including their %domain and 
range\footnote{See RDF schema in \url{https://www.w3.org/TR/rdf-primer/\#properties}}. The range indicates that the objects of a particular relation are instances of a designated class. %The domain is used to specify that a particular relation applies to a designated class. 
For example, %the domain of the \textit{PersonInstrument} relation is the class \textit{Person}, and 
the range of the \textit{PersonInstrument} relation is the class \textit{MusicalInstrument}.

\subsection{SATORI: Seek and entail for object prediction}
Recognizing textual entailment (RTE) is central in SATORI to validate triples. RTE is the task to determine whether a hypothesis H is true given a premise P. Typically for an input pair P and H, an RTE model assigns the labels Entailment (true), Contradiction (false) or Neutral \cite{maccartney-manning-2008-modeling}. We consider as hypothesis a natural language description of the candidate triple that we are validating, and as premises relevant text retrieved from the web. Thus SATORI has three major steps (see Fig. \ref{fig1}): i) retrieving premises, ii) getting candidate objects to generate new triples, and iii) validating triples using RTE. Along this section we use the input pair $s=John Lennon$, $r=PersonInstrument$ as an example for the object prediction task.

%To retrieve text passages that serve as premises SATORI queries a mainstream web search engine. Next, to get candidate objects from language models we use prompts. In addition, SATORI supports other sources of candidate objects include existing knowledge graphs or named entities extracted from text passages. Then, triples are generated using the input pair $(s,r)$ and the candidate objects. Finally, to validate triples SATORI first generates a hypothesis by transforming the candidate triple $(s,r,o)$ into natural language and then verifies whether the hypothesis can be inferred from the premises using RTE. 

\textbf{Retrieving premises}. The goal is to query a search engine using the input pair $s$ and $r$ to retrieve relevant text passages (featured snippets). For a relation $r$ we define a search template $t_{search,r}$ as a keyword-based query including a placeholder for the subject. The subject $s$ is replaced in the template $t_{search,r}$ and the output query is sent to the search engine API. We retrieve the top k featured snippets to use them as premises. For the example input pair and template $t_{search,r}$ \textit{= X plays instrument}, our query is \textit{John Lennon plays instrument}.

\textbf{Getting candidate objects}. Our first source of candidate objects is a pretrained language model. For each relation we define a template $t_{lm,r}$ of a prompt for the language model. The prompt is obtained by replacing the subject $s$ in the template $t_{lm,r}$. For example the template $t_{lm,r}$ \textit{= X plays $<$MASK$>$} becomes the prompt \textit{John Lennon plays $<$MASK$>$}. 
In a bidirectional transformer like BERT the output vectors corresponding to the mask tokens are fed into an output softmax layer over the vocabulary. Hence the language model output for a mask token is the words in the vocabulary and the score indicating the probability of being the mask token in the input sequence. We use a per-relation threshold  $T_{lm,r}$ on the word score. Predicted words with a score above the threshold are added to the set of candidate objects $objs=\{o_{i}\}$. Given an input pair $(s,r)$ and each candidate object $o_{i}$ we create a candidate triple $(s,r,o_{i})$

In addition, we consider existing knowledge graphs as source of candidate objects. %that match the range of the relation of interest. 
We use instances of the classes defined in the relation range as candidate objects. To retrieve such instances we use SPARQL queries.
First, we get the classes $C_j$ in the relation range from the schema of the knowledge graph that we are populating. Next, for each class in the relation range we send the following SPARQL query to the knowledge graph from where we want to obtain candidates:  
\textit{SELECT ?y WHERE ?y rdf:type $C_j$}.\footnote{While rdf:type is the standard property used to state that a resource is an instance of a class, some knowledge graphs could use other ad-hoc property. %For example in wikidata property P31 is used instead of rdf:type
} The retrieved entities are added to the set of candidate objects $objs=\{o_{i}\}$. For example, the relation \textit{personInstrument} expects entities of class \textit{MusicalInstrument} in the range. Thus, the SPARQL query generated for the example input pair is \textit{SELECT ?y WHERE ?y rdf:type MusicalInstrument}.

Both language models and knowledge graphs can generate a high number of non-relevant candidate objects. We use heuristics to filter out unrelated objects. The most basic filter is a stop word list including punctuation marks. Language models are known for assigning high probability to punctuation marks and articles. In addition, we filter out objects which are not explicitly mentioned in the text passages gathered in the premise retrieval stage. 

Named Entity Recognition NER can also be used to identify candidate objects for some relations depending on the classes in the relation range. Standard NER recognizes People, Locations, and Organizations\footnote{Due to the diverse nature of the MISC category we do not consider it.}. We apply NER on the texts passages retrieved from the web for the input pair. If there is a match between the classes in the relation range and the classes of the recognized entities then we add such entities to the set of candidate objects $objs=\{o_{i}\}$.  

\textbf{Validating triples}. We generate a short text description of a candidate triple $(s,r,o_{i})$ to be used as hypothesis to validate against the premises through RTE. We define per-relation templates $t_{h,r}$ to generate hypotheses containing placeholders for the subject (X) and object (Y). Then we replace the subject (X) and object (Y) placeholders with $s$ and $o_{i}$ to generate a hypothesis $H_{l}$. Given the candidate triple \textit{(John Lennon, PersonInstrument, Guitar)}, and template $t_{h,r}$ \textit{= {X} plays {Y}}, the generated hypothesis is $H_{l}$ \textit{= John Lennon plays guitar}. Next, we use a language model fine-tuned for the RTE task to evaluate whether $H_{l}$ is entailed on average by the $k$ premises. The objects $o_{i}$ corresponding to an entailed hypothesis are the final objects for the input tuple $(s,r)$. 

Language models fine-tuned for RTE address the task as a multi-class classification of premise and hypothesis pairs into Entailment, Contradiction and Neutral classes. For transformers like BERT the input is a sequence \textit{[CLS] premise [SEP] hypothesis [SEP]}. The [CLS] output vector $C \in R^{H}$, where H is the hidden size in the transformer, acts as the aggregated representation of the input sequence, and is connected to an output layer for classification $W\in^{K \times H}$, where K is the number of classes. Softmax is the activation function of the output layer and cross-entropy the loss function. In SATORI we focus on the Entailment and Contradiction classes, and softmax is applied only to the corresponding outputs. For each pair of premise and hypothesis the classifier generates scores for each of the classes. We use a per-relation threshold $T_{e,r}$ on the Entailment class score to accept the prediction as valid. 

\section{Evaluation setup}

\subsection{Dataset} 
The datasets used in the TAC KBP \cite{getman_laying_2018} evaluation series are tightly coupled with the text corpus used to mine facts: only information extracted from the corpus is considered valid to populate the knowledge graph. We discard such datasets since SATORI's web-based approach does not rely on a particular corpus. In addition, the evaluation of the slot-filling task, which is the most closely related to object prediction, is carried out manually. 

Other triple-based datasets like LAMA \cite{petroni_language_2019},  %used to analyze knowledge in language models, or 
FB15K-237 \cite{toutanova_representing_2015} and WN18 \cite{bordes_translating_2013}, %used for relation prediction in KBC, 
and their derived versions, were also discarded since they do not guarantee the completeness of objects for subject and relation pairs. Completeness of objects is important to evaluate precision and recall in object prediction. For instance, when predicting  objects for relation \textit{PersonInstrument}  we want to predict all the instruments and not only some instruments for a given individual. In LAMA triples are randomly sampled, while in FB15K-237 and WN18 only frequent entities and relations are selected. 

Moreover, optional relations that do not apply to every subject in the domain play an important role in object prediction evaluation. For example, the optional \textit{PlaceOfDeath} relation only applies to people that have passed away, not to all the instances of people. Thus, to evaluate whether an object prediction system must produce or not an object for a subject and relation pair, the dataset needs to include pairs for which an object is not expected. 

Therefore, we resort to the recently introduced LM\_KBC22~\cite{singhania_lm-kbc_2022}  dataset\footnote{https://lm-kbc.github.io/2022/} that includes all expected objects for a given subject and relation pair, is not tied to a particular corpus, and comprises subject-relation pairs for which objects are not expected. The dataset includes 12 relations exemplified by a set of subjects and a complete list of ground-truth objects per subject and relation pair. Five relations also contain subjects without any correct ground truth objects. %Cardinality of each relation range is presented in Table \ref{tab:lmkbc22}. 
In this dataset relation domains are diverse, including \textit{Person}, \textit{Organization}, or \textit{Chemical compound}. Similarly the range of relations includes \textit{Country}, \textit{Language}, \textit{Chemical element}, \textit{Profession}, \textit{Musical Instruments}, or \textit{Companies} to name a few classes. The training set includes 100 subjects per relation, and development and test sets each contain 50 subjects per relation. 

\begin{comment}
\begin{table}[t!]
  \centering
  \footnotesize
  \caption{Relations in LM\_KBC22, and min. and max. cardinality in their range.}
  \resizebox{0.6\linewidth}{!}{    
   \begin{tabular}{llll}
    \toprule
    \textbf{Relation} & \textbf{Train} & \textbf{Dev} & \textbf{Test} \\
    \midrule
    CountryBordersWithCountry & [0,17] & [0,14] & [0,11] \\
    CountryOfficialLanguage & [1,4] & [1,15] & [1,11] \\
    StateSharesBorderState & [1,14] & [1,15] & [1,14] \\
    RiverBasinsCountry & [1,6] & [1,10] & [1,9] \\
    ChemicalCompoundElement & [2,6] & [2,6] & [2,6] \\
    PersonLanguage & [1,6] & [1,5] & [1,7] \\
    PersonProfession & [1,23] & [1,19] & [1,20] \\
    PersonInstrument & [0,7] & [0,14] & [0,7] \\
    PersonEmployer & [1,8] & [1,8] & [1,8] \\
    PersonPlaceOfDeath & [0,1] & [0,1] & [0,1] \\    
    PersonCauseOfDeath & [0,3] & [0,2] & [0,2] \\
    CompanyParentOrganization & [0,5] & [0,1] & [0,3] \\
    \bottomrule
    \end{tabular}%
   }
  \label{tab:lmkbc22}%
\end{table}%
\end{comment}

\subsection{Metrics} 
KBP systems are required to make accurate decisions with good coverage of the predicted entities. Rank-aware metrics customary used in KBC such as Mean Rank, and Mean Recriprocal Rank %, that indicates how high in the ranking of predicted objects is the right prediction, 
are of little use in a real KBP setting. They indicate how high in the ranking of predicted objects are the right ones. However in KBP we are interested in making actual predictions. Therefore, we use standard classification metrics that are good indicators of system accuracy and coverage: precision, recall, and F1. 

\subsection{SATORI configuration}\label{configuration} We use the duckduckgo.com search engine to gather premises from the Web for each subject-relation pair. We leverage its python library\footnote{https://pypi.org/project/duckduckgo-search/} to send  queries to the text search service. This service returns the web page title, url, and featured snippet that we keep as a premise. The duckduckgo library is very convenient since it can be used straightaway in Python programs and does not require registration in a cloud-based platform, unlike with leading search engines.  We set k = 3 to gather at least three different premises that we can use to evaluate the hypotheses that we generate.

We evaluate three sources of candidate objects: pretrained language model (LM), knowledge graph (KG) and NER. We use BERT large cased as pretrained LM since it allows us to compare SATORI with the baseline model in the LM-KBC22 dataset. KG and NER are used together since  NER recognizes a limited number of entity types and some entity types might not be covered in a KG. To perform NER we use a transformer model fine-tuned on the task\footnote{\url{https://spacy.io/models/en\#en\_core\_web\_trf}}. We use NER to get candidate locations for relations \textit{StateSharesBorderState} and \textit{PersonPlaceOfDeath}, and candidate organizations for relations \textit{CompanyParentOrganization} and \textit{PersonEmployer}. We use Wikidata \cite{vrandevcic2014wikidata} as KG for the rest of the relations in the dataset. Wikidata includes instances of all classes in the ranges of the remaining relations: \textit{Language}, \textit{Country}, \textit{ChemicalElement}, \textit{MusicalInstrument}, \textit{Profession}, and \textit{CauseOfDeath}.

To validate  candidate triples we test three language models fine-tuned for the entailment task. We choose the models according to their performance on the entailment task and the number of parameters in the model. We use a DeBERTa Extra Large (XL) model \footnote{\url{https://huggingface.co/microsoft/deberta-v2-xlarge-mnli}} with 900M parameters that reports 91.7 accuracy on the MNLI dataset \cite{williams-etal-2018-broad}. We also use DeBERTa Extra Small (XS)\footnote{\url{https://huggingface.co/microsoft/deberta-v3-xsmall}} model with 22M parameters that reports 88.1 accuracy on the same dataset. The final model that we test is a BERT large model\footnote{\url{https://huggingface.co/boychaboy/MNLI\_bert-large-cased}} with 336M parameters that reports 86.7 accuracy on MNLI.

Templates $t_{lm,r}$, $t_{search,r}$, and $t_{h,r}$  used in the experiments are listed in the paper repository.\footnote{\url{https://github.com/satori2023/Textual-Entailment-for-Effective-Triple-Validation-in-Object-Prediction}} Particularly, templates $t_{lm,r}$ are reused from the language model baseline that we describe below. %Finally, we use  the following notation to distinguish SATORI configurations: 

%SATORI\_\textless Object\_Source\textgreater \_\textless Entailment\_Model\textgreater.

\subsection{Baselines}\label{baselines}
The first baseline that we use is prompting a language model. Such baseline allows to test our hypothesis that triple validation through textual entailment can benefit object prediction from language models. In addition, we are interested in comparing the knowledge in language models with other sources of knowledge including knowledge graphs and text passages related to the input subject and relation pair. Thus, we use as baselines a state of the art extractive question answering model and a relation extraction model repurposed for object prediction. 

\textbf{LM-baseline} We reuse the baseline system from the LM\_KBC challenge~\cite{singhania_lm-kbc_2022}, which prompts a BERT-large-cased model. We use the templates to transform triples into prompts made available in the challenge repository.\footnote{https://github.com/lm-kbc/dataset} The baseline uses a threshold $T_{lm,r}$ on the likelihood of words to fill in the mask token. Stop words are also filtered out. In addition, following Li et al. \cite{Li2022TaskspecificPA} we further-pretrain the language model on the masked language modeling objective using training data from the LM\_KBC22 dataset. We transform  triples in the training dataset into prompts where the objects are masked, and train the model to predict the right objects (see section \ref{regimes} for further pretraining details). 

The BERT-based system that scored highest in the LM\_KBC22 challenge trains several BERT models (one model per relation) and further pretrains the models with data from Wikidata \cite{Li2022TaskspecificPA}. While we could have used such approach as a baseline and integrated it in SATORI as source of candidate objects, we decided against it since we think such approach does not scale due to the number of models being trained. 

\textbf{QA-baseline} We use an extractive Question Answering (QA) system as baseline. To this end, we transform each subject-relation pair using per-relation templates into a question, where the answer is the expected object. The QA system then attempts to extract the answer from the text passages that we gather from the Web as premises in SATORI for the same input pair (see section \ref{SATORI} for the premise retrieval procedure). QA per-relation templates $t_{qa,r}$ have a placeholder ($X$) for the subject entity and follow a similar basic pattern with minor variations: \textit{What relation*} $X$?, where relation* is a text fragment derived from the relation label. For example, the template for the \textit{PlaysInstrument} relation is: \textit{What instruments plays X?}. The complete set of QA templates is available in the paper repository. %As text passages we use the premises that we retrieved from the Web in SATORI (See section \ref{SATORI} for the premise retrieval procedure). 

As QA model we use DeBERTa large\footnote{https://huggingface.co/deepset/deberta-v3-large-squad2} fine-tuned on SQUAD 2.0. DeBERTa is the highest scoring single model in the SQuAD 2.0 leaderboard\footnote{https://rajpurkar.github.io/SQuAD-explorer/} (F1=90.3) that is available on hugginface.co. We slightly post-process DeBERTa's answers containing lists of items to extract each item. For example, for the question \textit{What instruments plays John Lennon?}, DeBERTa extracts the answer \textit{guitar, keyboard, harmonica and horn}. From such list we extract each single instrument as a possible answer. Along with the answer span DeBERTa generates a score that indicates the probability of the answer. We use a per-relation threshold $T_{qa,r}$ on that score to accept or reject an answer. In addition we further fine-tune the QA model using questions and answers derived from the training set and the premise retrieved in SATORI as text passages from where the answers can be extracted (see section \ref{regimes} for further details on QA fine-tuning). 

%\hl{We only use the QA baseline in the zero-shot scenario. Generating the training instances for the extractive QA model in few-shot regimes is out of the scope of this paper. QA training instances requires gold text passages that contains the complete list of objects for a given input subject and relation pair, and there is not guarantee to find such passages in the Web.. In addition, even though we can find a text passage with all the answers for a given question, the answers can be scattered in the text passage, and not necessarily arranged in a list. Extractive QA models trained on SQuAD only consider contiguous text span as answers.}

\textbf{RE-baseline} We also test the  state of the art relation extraction system REBEL \cite{huguet-cabot-navigli-2021-rebel-relation} in the object prediction task. REBEL is a sequence to sequence model based on BART\cite{lewis-etal-2020-bart} that performs end-to-end relation extraction for more than 200 different relation types. REBEL  autoregressively generates each triple present in the input text. To use REBEL we need to map the relations it supports with relations in the LM\_KBC22 dataset. From the 12 relations in LM\_KBC22 we find 10 relations with the same or very similar semantics in REBEL.\footnote{The relation mapping can be found in the paper repository} For the \textit{ChemicalCompoundElement}, one of the remaining relations, we establish a mapping with the broader \textit{has part} relation. The \textit{PersonCauseOfDeath} relation could not be mapped. We use REBEL to generate triples from the text passages previously retrieved as premises for input subject-relation pairs. Finally, we extract the objects from those triples as the predicted objects.

\subsection{Pretrained models and training regimes}\label{regimes} 
We evaluate pretrained models off the shelf and once they are further trained using the LM\_KBC22 dataset. Since the test set is withheld by the dataset authors, we test on the development set and use a held-out 20\% of the training set as a new development set, leaving the remaining 80\% as training data. 

First we evaluate SATORI and the baselines using available pretrained models (see sections \ref{configuration} and \ref{baselines}). To adjust the parameters that control the predictions of the different models for each relation, i.e. $T_{lm,r}$, $T_{e,r}$ in SATORI, $T_{lm,r}$ in the LM\_baseline and $T_{qa,r}$ in the QA\_baseline, we search for the best thresholds F1-wise in the range 0.01 to 0.99 with 0.01 increments over the union of training and development sets.

In addition, we train the models that we use in SATORI and the baselines on random samples of 5\%, 10\% and 20\% of the training set  corresponding on average to 4, 8, and 16 examples per relation. We also use 100\% of the training data.  
The same percentages are used to sample the development set in each training scenario. %In the development set each subject and relation pair has the whole list of possible objects. 
Each training scenario is repeated 10 times, and evaluation metrics are averaged. We follow the same strategy to adjust the parameter that we use for pretrained models using the training and development sets. In the following we describe how we train the models.

\textbf{Language model}. The language model used in SATORI and the LM-baseline (BERT large) is further trained on the masked language model objective using the data from the training set in each scenario. We transform triples in the training and development set into prompts using the per-relation templates $t_{lm,r}$ and train the model to fill in the mask tokens with the corresponding objects. For the masked language modeling training objective, we set the hyper-parameters following \cite{Li2022TaskspecificPA}. That is, we train the model for 10 epochs, use a learning rate of 5e-6, an a batch size of 32. We save checkpoints at the end of every epoch, considering as the best checkpoint the one with the lowest development set loss.

\textbf{Entailment model}. %Similarly, we further fine-tune the entailment model in each training scenario. 
An entailment training instance comprises a premise, hypothesis, and a label indicating Entailment or Contradiction. We create entailment training instances as follows. For each triple in the training set we use the text passages retrieved as premises for the subject and relation in the triple. If the subject and object are mentioned in the text passage we generate a positive entailment example using the passage as premise and the hypothesis that we generate using the corresponding template $t_{h,r}$. To generate negative examples, i.e., contradictions, we replace the object in the positive example with an incorrect object, and use a premise retrieved for the input pair where such object is mentioned. To obtain the incorrect object we prompt the language model using the per-relation template $t_{lm,r}$ where we replace the input subject. We keep  objects appearing in any of the premises for the input pair, that are not related to the input subject-relation pair in the training set. If we do not find any object from the language model that satisfies the previous condition, we look for incorrect objects in the training set for the same relation. 

To train the entailment model, we reuse the classification scripts available in HuggingFace.\footnote{https://github.com/huggingface/transformers/tree/v4.24.0/examples/pytorch/text-classification} We use the default hyper-parameters: 3 epochs, learning rate of 2e-5, and maximum sequence length of 128. Due to hardware limitations we use a batch size of 8, and gradient accumulation steps of 4. Particularly, to fine-tune the DeBERTa xlarge model, we reduce further the batch size to 1 and increase gradient accumulation steps to 32, applying gradient checkpointing, and use "Adafactor" optimizer instead of the default "AdamW".

\textbf{QA-baseline}. %We also further train the QA model that we use as baseline. 
A training instance for an extractive QA model includes a question, an answer and a text from where the answer is extracted. To generate training instances we first pre-process the training set and obtain for each subject and relation the complete lists of objects. Next, we get the text passages retrieved as premises for each subject and relation pair. For entries with empty object lists, we use the first retrieved passage and also use the empty gold answer. 

If there is only one object for the input subject and relation pair, and there is a text passage where the object appears, we generate a training example with the passage, the question that we generate using the template $t_{qa,r}$ and the object as answer. If there is more than one object for the subject and relation the only way to extract them from a passage using a QA model is if they are arranged as a list of terms in a contiguous text span. Therefore we look for text passages containing every object with the condition that they must be located at most three tokens away from each other. We select the passage with the span containing the highest number of objects, and generate the training example with such passage, the question that we generate using the template $t_{qa,r}$, and the text span where objects appear as answer.

To train the question answering model we leverage HuggingFace scripts\footnote{https://github.com/huggingface/transformers/tree/v4.24.0/examples/pytorch/question-answering}, and use the default hyper-parameters: batch size of 12, learning rate of 3e-5, 2 training epochs, and a maximum sequence length of 384.

\textbf{RE-baseline.} Training instances for REBEL consist of a short text, followed by the triples that can be extracted from that text. Triples are described in a REBEL-specific format using special tokens. Relations in triples are indicated using their text descriptions. To transform instances from the LB\_KBC22 dataset into REBEL training instances we first obtain for each subject and relation the complete lists of objects. Next, we get the text passages retrieved as premises for each  subject and relation pair. If we find a text passage containing some of the objects related to subject and relation pair we create a REBEL training instance with such text, and the triples for the subject, relation and objects found in the text. To train the model we use the script\footnote{https://github.com/Babelscape/rebel/blob/main/src/train.py} provided by the REBEL authors. We use the default training parameters in the script.

\section{Evaluation results} \label{evaluation}
Table \ref{tab:results} shows evaluation results in the object prediction task for SATORI and  baselines using pretrained models and models further trained on the different training regimes. Note that SATORI is evaluated using different RTE models for triple validation and also using different sources of candidate objects. 

\subsection{Language model prompting and triple validation} Let us start with  Q1 (does candidate triple validation through textual entailment improve object prediction results over pretrained language model prompting?). We compare SATORI using RTE for triple validation and a pretrained language model as source of candidate objects (Table \ref{tab:results}: first three rows in the Pretrained column) against the LM\_baseline using only a pretrained language model. 

Results show that in such scenario triple validation using any RTE model improves the LM\_baseline in all evaluation metrics. The largest gain is in precision, with an improvement that ranges from 14 to 15.5 points depending on the RTE model. Recall is also improved, although to a lesser extent, in the range of  1.9 to 3.2. Improvement in precision shows that triple validation using RTE is effective to filter out non-relevant objects predicted by the language model. Since the system is more precise, the threshold limiting the number of objects predicted per relation $T_{lm,r}$ is actually lower in SATORI than in the LM\_baseline. A lower $T_{lm,r}$ allows obtaining more candidates from the language model, thus the improvement on recall. Overall, triple validation using RTE improves over the LM\_baseline with a margin ranging between 4.7 and 5.4 F1 points and research question Q1 is therefore satisfactorily addressed. 

\begin{table}[ht!]
  \centering
  \caption{\small Object prediction using pretrained models vs. such models additionally trained in different scenarios. The models include the LM from where we get objects and the entailment model that we use for triple validation. The SATORI versions evaluated include three sources of candidate objects (Language Model LM, Knowledge Graph KG, Named Entities NER) and three entailment models (DeBERTa xs, BERT large, DeBERTa xl).}
 \resizebox{\textwidth}{!}{  
    \begin{tabular}{ccrrrrrrrrrrrrrrr}
    \toprule
    \multicolumn{2}{c}{\multirow{2}[3]{*}{\textbf{SATORI}}} & \multicolumn{3}{c}{\multirow{2}[3]{*}{\textbf{Pretrained}}} & \multicolumn{12}{c}{\textbf{Training Data}} \\    
\multicolumn{2}{c}{} & \multicolumn{3}{c}{}  & \multicolumn{3}{c}{5\%} & \multicolumn{3}{c}{10\%} & \multicolumn{3}{c}{20\%} & \multicolumn{3}{c}{100\%} \\
\cmidrule(ll){1-2}
\cmidrule(ll){3-5}
\cmidrule(ll){6-17}
    \multicolumn{1}{c}{\textbf{RTE Model}} & \multicolumn{1}{c}{\textbf{Obj. Source}} & \multicolumn{1}{c}{P} & \multicolumn{1}{c}{R} & \multicolumn{1}{c}{F} & \multicolumn{1}{c}{P} & \multicolumn{1}{c}{R} & \multicolumn{1}{c}{F} & \multicolumn{1}{c}{P} & \multicolumn{1}{c}{R} & \multicolumn{1}{c}{F} & \multicolumn{1}{c}{P} & \multicolumn{1}{c}{R} & \multicolumn{1}{c}{F} & \multicolumn{1}{c}{P} & \multicolumn{1}{c}{R} & \multicolumn{1}{c}{F} \\
\cmidrule(ll){1-2}
\cmidrule(ll){3-5}
\cmidrule(ll){6-17}
    \multicolumn{1}{l}{DeBERTa xs} & \multicolumn{1}{l}{LM} & 83.6  & 48.6  & 47.9  & 80.5  & 51.7  & 47.9  & 81.9  & 51.5  & 48.4  & 84.1  & 51.3  & 50.3  & \textbf{91.3} & 44.7  & 46.8 \\
    \multicolumn{1}{l}{BERT large} & \multicolumn{1}{l}{LM} & 82.1  & 49.9  & 48.6  & 79.9  & 52.2  & 48.1  & 81.7  & 51.8  & 48.7  & 84.8  & 51.2  & 50.6  & 89.3  & 46.0  & 47.1 \\
    \multicolumn{1}{l}{DeBERTa xl} & \multicolumn{1}{l}{LM} & 83.0  & 49.5  & 48.5  & 82.4  & 51.9  & 48.5  & \textbf{84.3} & 52.1  & 50.1  & \textbf{87.0} & 52.0  & 51.4  & 91.1  & 47.8  & 48.4 \\
    \multicolumn{1}{l}{DeBERTa xs} & \multicolumn{1}{l}{KG+NER} & 67.7  & 61.8  & 55.2  & 67.0  & 60.9  & 52.1  & 69.5  & 60.1  & 52.6  & 71.0  & 60.8  & 53.8  & 73.5  & 60.4  & 54.7 \\
    \multicolumn{1}{l}{BERT large} & \multicolumn{1}{l}{KG+NER} & 67.0  & 62.9  & 55.2  & 68.2  & 61.8  & 52.2  & 70.0  & 61.1  & 52.9  & 69.9  & 63.0  & 54.9  & 66.9  & 64.8  & 55.1 \\
    \multicolumn{1}{l}{DeBERTa xl} & \multicolumn{1}{l}{KG+NER} & 70.0  & 62.5  & \textbf{55.9} & 71.4  & 62.2  & \textbf{54.9} & 72.6  & 62.3  & \textbf{55.1} & 70.7  & 63.6  & \textbf{56.5} & 68.3  & 64.0  & 54.8 \\
    \multicolumn{1}{l}{DeBERTa xs} & \multicolumn{1}{l}{LM+KG+NER} & 66.6  & 62.0  & 54.4  & 65.4  & 61.2  & 51.5  & 67.1  & 60.8  & 51.8  & 69.1  & 61.5  & 53.4  & 73.3  & 60.5  & 54.6 \\
    \multicolumn{1}{l}{BERT large} & \multicolumn{1}{l}{LM+KG+NER} & 64.5  & 63.2  & 54.7  & 65.7  & \textbf{62.6} & 51.5  & 68.2  & 61.6  & 52.7  & 68.6  & 63.5  & 54.8  & 67.2  & \textbf{65.1} & 55.7 \\
    \multicolumn{1}{l}{DeBERTa xl} & \multicolumn{1}{l}{LM+KG+NER} & 63.5  & \textbf{64.3} & 53.9  & 69.4  & 62.6  & 54.2  & 70.6  & \textbf{63.1} & 54.8  & 69.5  & \textbf{64.2} & \textbf{56.5}  & 70.0  & 64.3  & \textbf{56.6} \\
\cmidrule(ll){1-2}
\cmidrule(ll){3-5}
\cmidrule(ll){6-17}
    \multicolumn{2}{c}{LM-baseline} & 68.1  & 46.7  & 43.2  & 59.9  & 50.8  & 43.3  & 63.6  & 49.0  & 44.3  & 67.2  & 49.6  & 46.9  & 52.2  & 48.1  & 41.4 \\
    \multicolumn{2}{c}{QA-baseline} & 67.1  & 42.3  & 40.9  & 54.1  & 45.2  & 38.5  & 57.9  & 45.4  & 40.6  & 57.6  & 46.5  & 41.7  & 57.4  & 52.7  & 47.3 \\
    \multicolumn{2}{c}{RE-baseline} & \textbf{85.5} & 33.3  & 31.8  & \textbf{83.8} & 41.8  & 40.6  & 81.3  & 43.0  & 42.0  & 78.6  & 44.7  & 43.2  & 70.0  & 54.9  & 48.2 \\
    \bottomrule
    \end{tabular}%
    }
  \label{tab:results}%
\end{table}%

\subsection{Additional pretraining and fine-tuning scenarios}
\begin{figure}[ht!]
        \centering
        \begin{subfigure}[b]{0.475\textwidth}
            \centering
            \includegraphics[width=\textwidth]{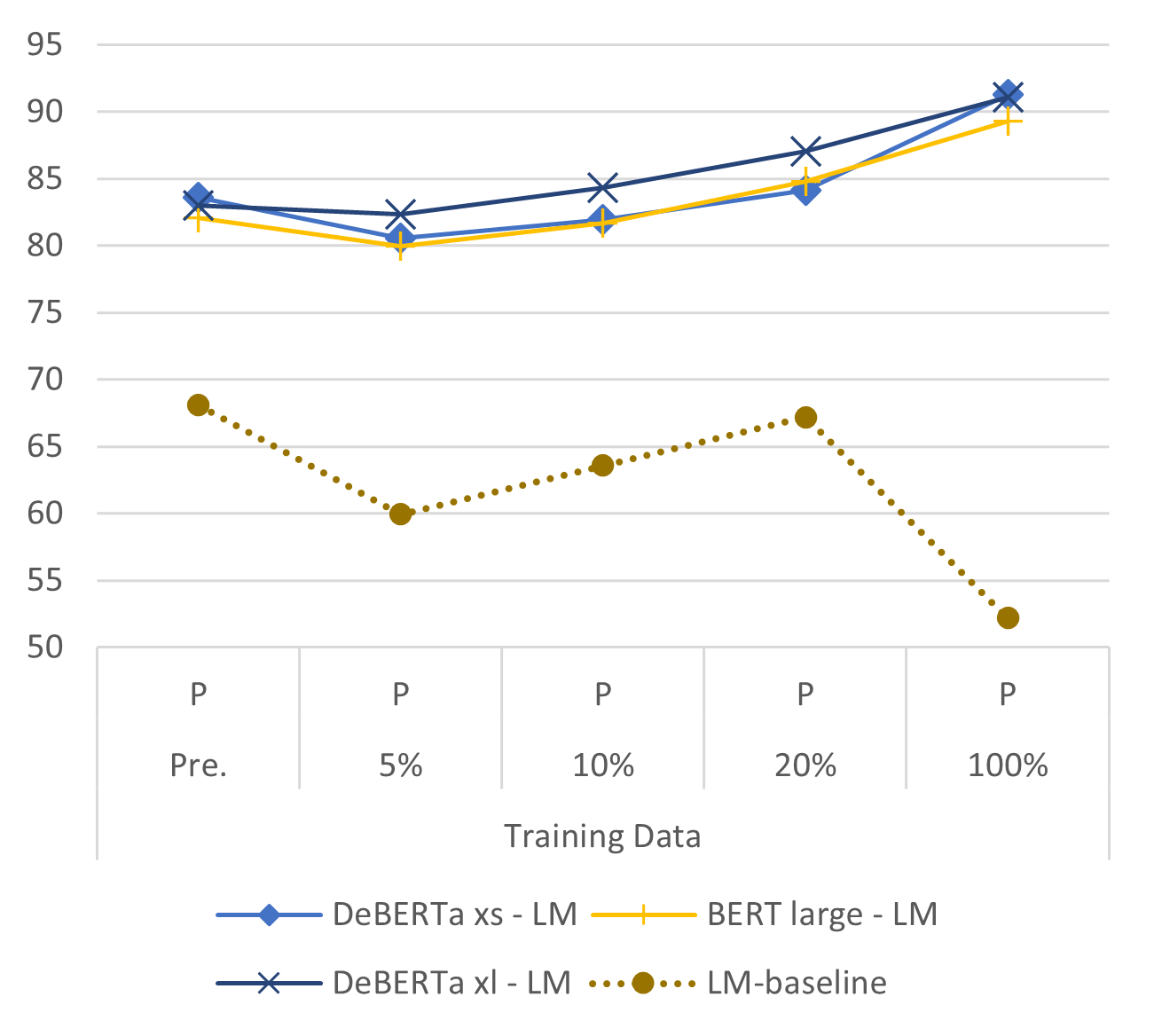}
            \caption[]%
            {{\footnotesize Precision:  LM baseline and SATORI using different RTE models and objects from LM.}}    
            \label{fig:evalP}
        \end{subfigure}
        \hfill
        \begin{subfigure}[b]{0.475\textwidth}  
            \centering 
            \includegraphics[width=\textwidth]{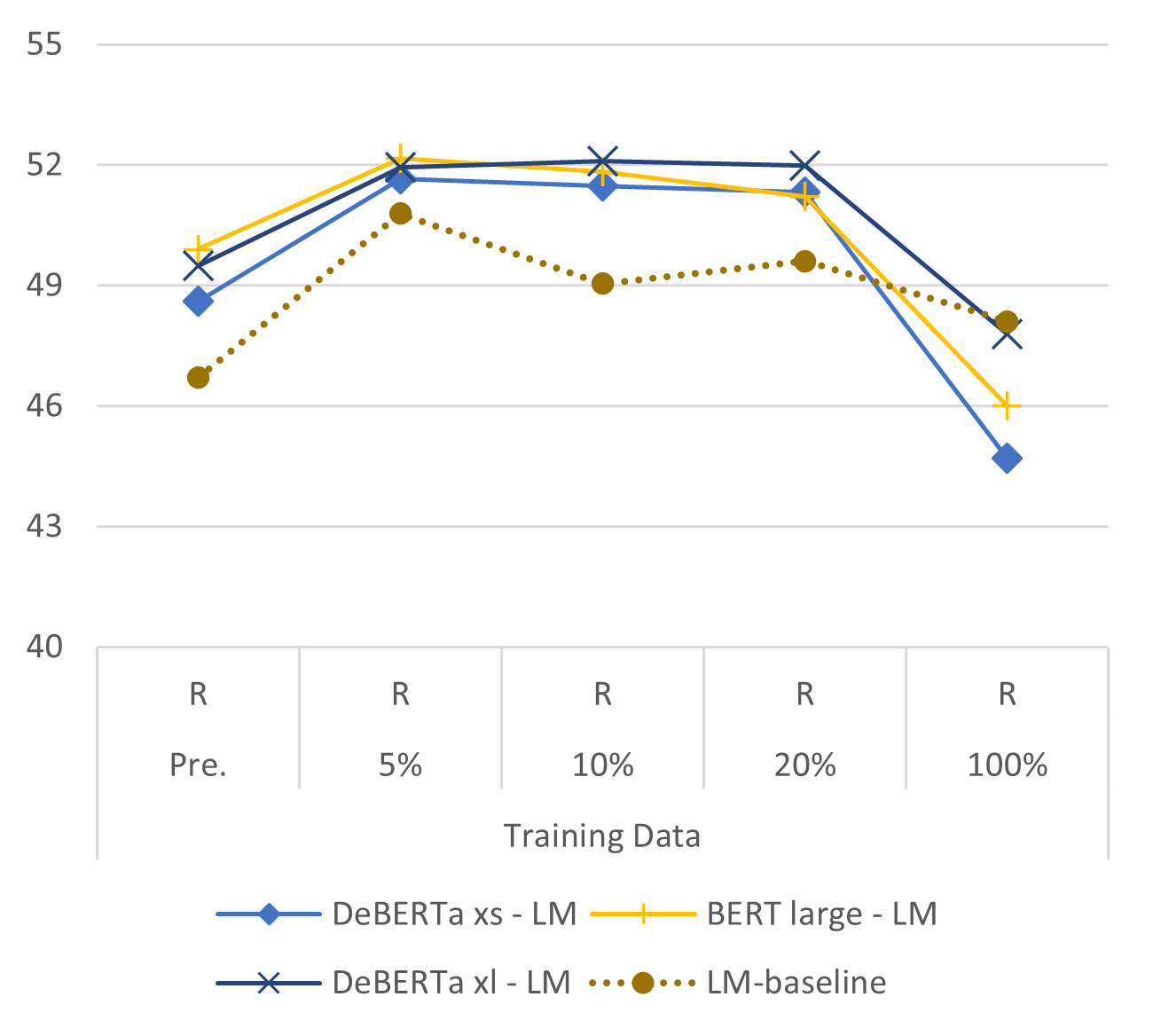}
            \caption[]%
            {{\footnotesize Recall: LM baseline and SATORI using different RTE models and objects from LM.}}    
            \label{fig:evalR}
        \end{subfigure}
        \vskip\baselineskip
        \begin{subfigure}[t]{0.475\textwidth}   
            \centering 
            \includegraphics[width=\textwidth]{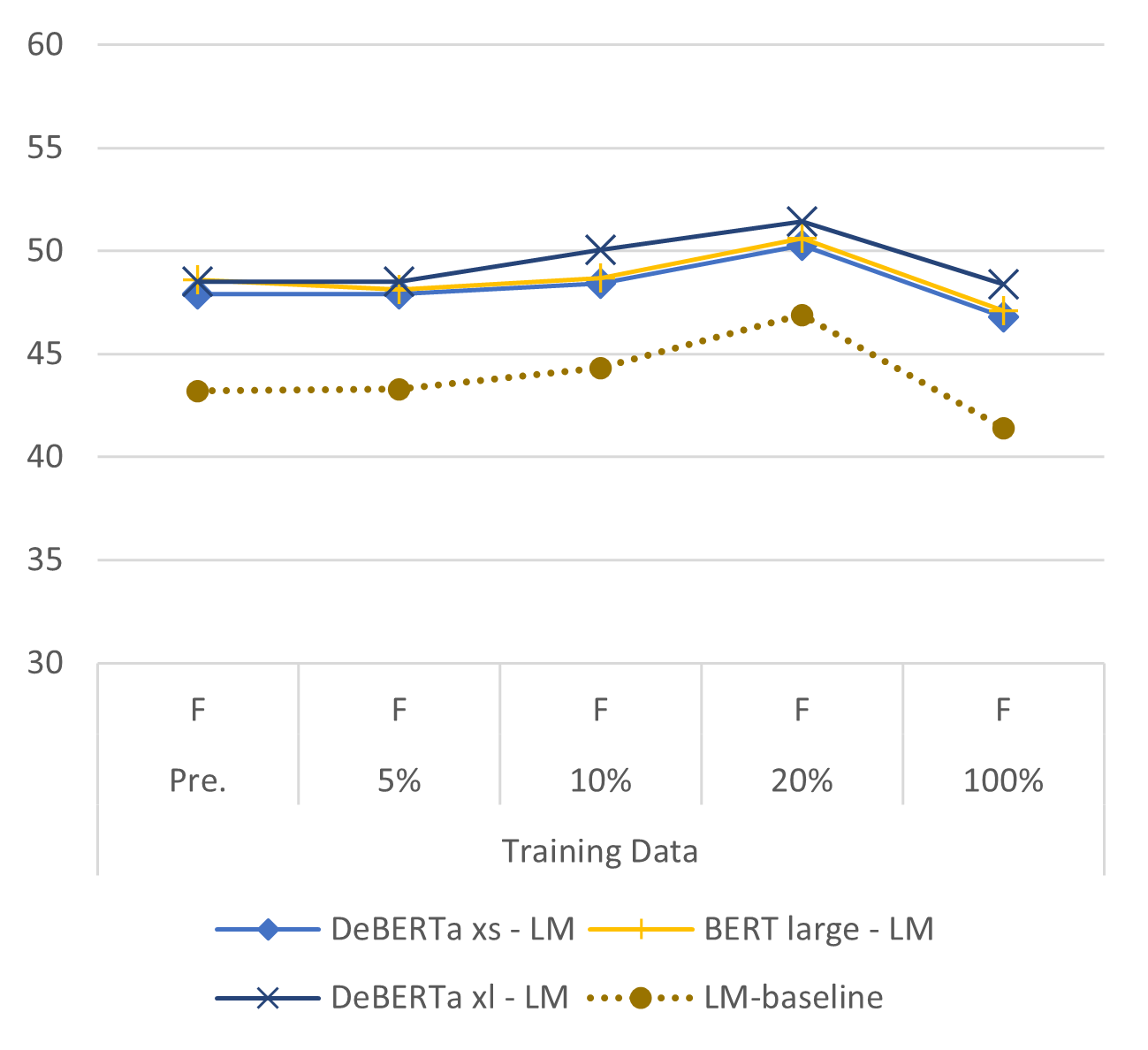}
            \caption[]%
            {{\footnotesize F1: LM baseline and SATORI using different RTE models and objects from LM}}    
            \label{fig:evalF1}
        \end{subfigure}
        \hfill
        \begin{subfigure}[t]{0.475\textwidth}   
            \centering 
            \includegraphics[width=\textwidth]{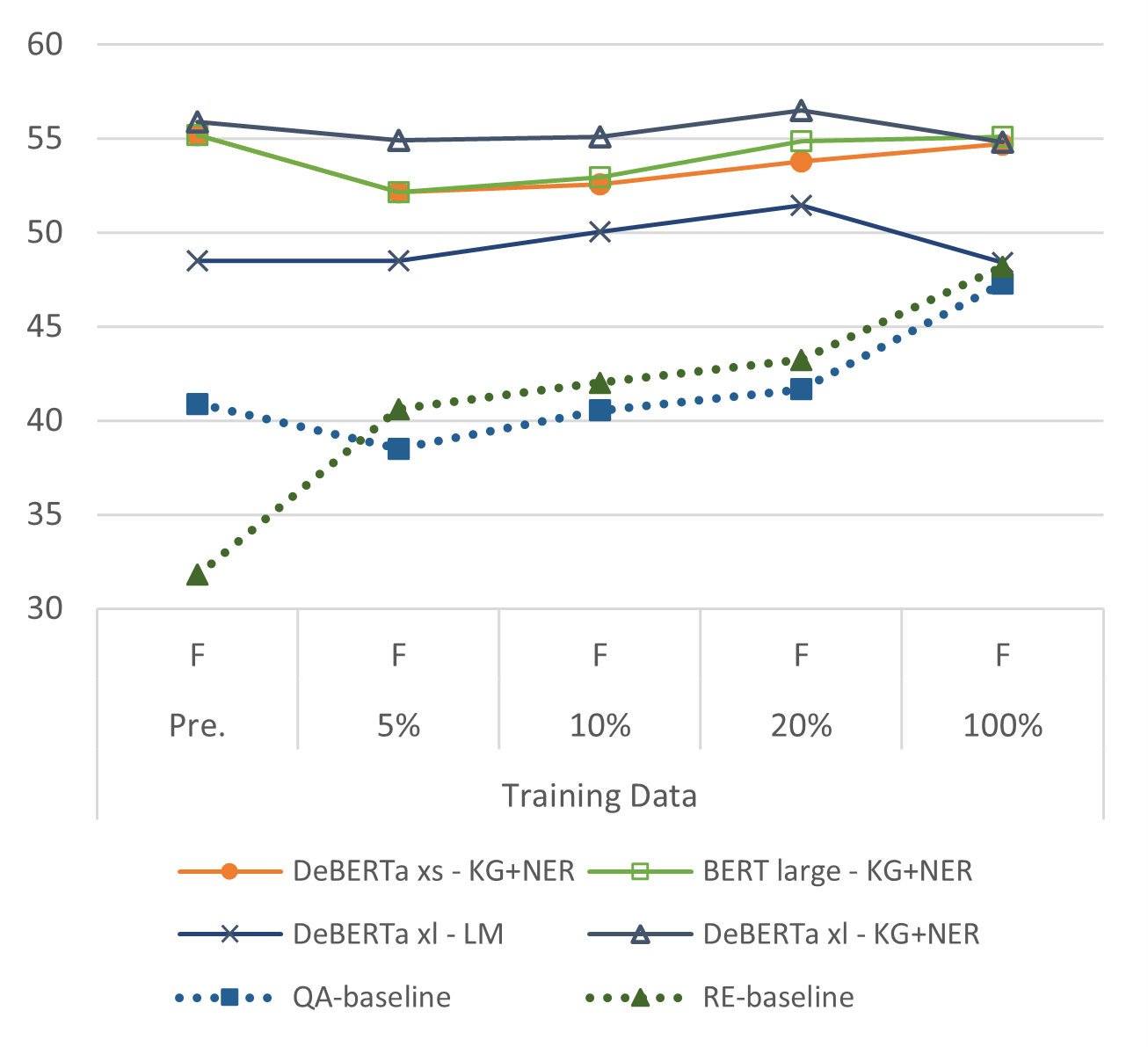}
            \caption[]%
            {{\footnotesize F1: best SATORI getting objects from LM, vs. using other knowledge sources and the QA and RE baselines.}}    
            \label{fig:evalF2}
        \end{subfigure}
        \caption[ ]
        {\small Object prediction evaluation on the LM\_KBC22 dataset.} 
        \label{fig:evalcharts}
\end{figure}

To address research question Q2 (how do further pretraining the language model and fine-tuning for entailment-based triple validation under different data regimes impact object prediction?) we analyse evaluation results when models are further trained. In Fig. \ref{fig:evalF1} we can see that triple validation using any RTE model improves F1 across all training scenarios compared to LM\_baseline. Interestingly the LM\_baseline F1 is lower in full training than the pretrained version and the other low data training regimes. Pending further experimentation, this can indicate a case of catastrophic forgetting \cite{mccloskey1989catastrophic,Goodfellow2013AnEI,serra2018overcoming}, where, upon further pretraining, language models might lose some of the (in this case, factual) knowledge previously acquired during pretraining.

The more training data available, the more precise the predicted objects using triple validation are (see Fig. \ref{fig:evalP}). Triple validation also improves recall, although in a narrower margin, across most training scenarios except in full training (see Fig. \ref{fig:evalR}). Nevertheless, the use of more training data does not imply an improvement in recall for SATORI. Such recall pattern reflects the higher precision the system achieves when more training data is used. That is, the triple validation filter becomes more and more strict, thus accepting less predicted objects. In low data regimes (5\% of the training data) the LM\_baseline precision decreases compared to results achieved with pretrained models. However, the precision  using triple validation does not fall at the same rate. This shows that the triple validation component is effective in low data regimes to filter out non relevant predicted objects. 

\subsection{Language models vs structured knowledge sources and information extraction methods}
To answer research question Q3 (how do language models compare to other potential sources of structured knowledge, as well as to methods based on extracting information from text, for the generation of candidate triples?) we first start comparing the evaluation results of LM\_baseline and the QA and RE baselines, which in addition leverage text passages. Table~\ref{tab:results} shows that in pretrained and most training regimes, except in full training, the QA and RE baselines achieve worse F1 than retrieving objects from the language models. Nevertheless, in a full training scenario the QA and RE baselines achieve an F1 score that is higher than the one obtained with LM\_baseline and similar to the one that we obtained with SATORI, using triple validation (see Fig. \ref{fig:evalF2}). Therefore, when enough training data is available extractive methods relying on state-of-the art QA and Relation Extraction are better for object prediction than relying only on LM as source of objects, and comparable to using triple validation to curate candidate objects retrieved from a LM. 

Nevertheless, the best performance in object prediction is achieved when we use KG and NER as source of candidate objects along with triple validation (see Fig. \ref{fig:evalF2}). Using KG and NER plus triple validation is the best option for object prediction in low data training regimes (5\%, 10\% and 20\%). Enhancing the list of candidate objects from KG and NER including objects from LM slightly enhances F1 in full training. Manual inspection shows that in full training, the number of candidate objects obtained from the LM is 794, from the KG it is 2338 and from NER it is 3506. While the number of objects from LM is lower due to the thresholds on the LM score and filters, such objects contribute to slightly increase the overall performance when combined with objects from KG and NER. The large number of objects obtained from KG+NER and the evaluation results shows that triple validation is effective to filter out non-relevant objects. In fact, if we remove the triple validation component and use objects from LM, NER and KG in the full training scenario we get 0.276 precision, 0.761 recall, and 0.327 F1. Such training scenario without triple validation is even worse than the LM\_baseline in terms of F1.

\section{Conclusions}
In this paper we posit that object prediction relying on prompting language models can be enhanced through triple validation using textual entailment. Our experiments show that triple validation provides a boost in object prediction performance using pretrained models and when such models are further fine-tuned under different training regimes. In addition, we compare language models with other sources of knowledge including existing knowledge graphs and relevant text passages where techniques such as named entity recognition NER, question answering QA and relation extraction RE can be applied for object prediction. 

We find that in low data regimes, getting objects using language model prompting  with or without triple validation is better than using extractive QA and RE models. However in full training the performance of the QA and RE models surpass language model prompting and reach the performance achieved when we validate the triples proposed by the language model. Moreover, using existing knowledge graphs and NER on text passages as source of candidate objects along with triple validation shows the overall best performance when pretrained models and models fine-tuned in low data regimes are evaluated. Adding candidate objects from language models to the ones found on KG and using NER is the best option in full training.  

The results presented in this paper are limited to the language model that we use, particularly BERT large, as primary source of candidate objects. Further research is needed to understand whether triple validation will continue having a positive effect in the object prediction task using a language model with a larger number of parameters and if this could balance the current gap between LMs and KGs when it comes to proposing candidate objects that we identified here. As future work, we think Entailment Graphs \cite{hosseini-etal-2021-open-domain,chen-etal-2022-entailment} could be useful to get better per-relation templates or to improve the entailment model used in triple validation. In addition, we plan to assess how triple classification \cite{jaradeh2021triple} compares to textual entailment for triple validation, and whether fact checking techniques \cite{gerber2015} applied to triple validation  brings new improvements. Finally, another research avenue is to use triple validation along with the QA and RE models given the good results that we obtained in full training when we use them for object prediction.

\paragraph*{Supplemental Material Statement:} All necessary resources to reproduce the experiments presented in section \ref{evaluation} are available in Github.\footnote{\url{https://github.com/expertailab/Textual-Entailment-for-Effective-Triple-Validation-in-Object-Prediction}} The repository includes code, templates and text passages that we use as premises in triple validation, and as input in the QA and RE baselines. 

\subsubsection*{Acknowledgements.} 
We are grateful to the European Commission (EU Horizon 2020 EXCELLENT SCIENCE - Research Infrastructure under grant agreement No. 101017501 RELIANCE) and ESA (Contract No. 4000135254/21/NL/GLC/kk FEPOSI) for the support received to carry out this research.
%
% ---- Bibliography ----
%
% BibTeX users should specify bibliography style 'splncs04'.
% References will then be sorted and formatted in the correct style.
%
\bibliographystyle{splncs04}
\bibliography{references}

\end{document}